\documentclass[a4paper]{article}

\usepackage{authblk}
\usepackage[utf8]{inputenc}
\usepackage[T1]{fontenc}
\usepackage{pslatex}
\usepackage[english]{babel}
\usepackage{fancyhdr}
\usepackage{hyperref}
\usepackage{amsmath,amssymb,amsfonts}
\usepackage{graphicx}
\usepackage{amsmath}
\usepackage{amsthm}
\usepackage{amsfonts}
\usepackage{mathtools}
\usepackage{amssymb}
\usepackage{algorithm}
\usepackage[noend]{algpseudocode}
\usepackage[noEnd=false]{algpseudocodex}
\usepackage{microtype}
\usepackage{rotating}
\usepackage{todonotes}
\usepackage{tikz}
\usepackage{tikz-qtree}
\usepackage{xparse}
\usepackage{appendix}
\usepackage{url}
\usepackage{tabularx}
\usepackage{paralist}
\usepackage{booktabs}
\usepackage{multirow}
\usepackage{subcaption}
\usepackage{cancel}
\usetikzlibrary{calc,shapes.callouts,shapes.arrows}
\usetikzlibrary{automata}
\usetikzlibrary{positioning}
\usetikzlibrary{decorations.pathreplacing}
\usetikzlibrary{decorations.pathmorphing}
\usetikzlibrary{arrows}
\usetikzlibrary{shapes}

\newcommand{\F}{\mathbb{F}}

\newtheorem*{problem*}{Problem}

\providecommand{\keywords}[1]{\textbf{\textit{Keywords }} #1}

\begin{document}

\title{Look into the Mirror: Evolving Self-Dual Bent Boolean Functions}

\author[1,2]{Claude Carlet}
\author[3]{Marko \DJ urasevic}
\author[3]{Domagoj Jakobovic}
\author[4]{Luca Mariot}
\author[5]{Stjepan Picek}

\affil[1]{{\normalsize Department of Mathematics, Universit\'{e} Paris 8, 2 rue de la libert\'{e}, 93526 Saint-Denis Cedex, France}}

\affil[2]{{\normalsize University of Bergen, Bergen, Norway} \\
	
	{\small \texttt{claude.carlet@gmail.com}}}

\affil[3]{{\normalsize Faculty of Electrical Engineering and Computing, University of Zagreb, Unska 3, Zagreb, Croatia} \\

{\small \texttt{marko.durasevic@fer.hr, domagoj.jakobovic@fer.hr}}}

\affil[4]{{\normalsize Semantics, Cybersecurity and Services Group, University of Twente, 7522 NB Enschede, The Netherlands} \\
	
	{\small \texttt{l.mariot@utwente.nl}}}

\affil[5]{{\normalsize Digital Security Group, Radboud University, Postbus 9010, 6500 GL Nijmegen, The Netherlands} \\
	
	{\small \texttt{stjepan.picek@ru.nl}}}
	
\maketitle

\begin{abstract}
Bent Boolean functions are important objects in cryptography and coding theory, and there are several general approaches for constructing such functions. Metaheuristics proved to be a strong choice as they can provide many bent functions, even when the size of the Boolean function is large (e.g., more than 20 inputs). While bent Boolean functions represent only a small part of all Boolean functions, there are several subclasses of bent functions providing specific properties and challenges. One of the most interesting subclasses comprises (anti-)self-dual bent Boolean functions. This paper provides a detailed experimentation with evolutionary algorithms with the goal of evolving (anti-)self-dual bent Boolean functions. We experiment with two encodings and two fitness functions to directly evolve self-dual bent Boolean functions. Our experiments consider Boolean functions with sizes of up to 16 inputs, and we successfully construct self-dual bent functions for each dimension. Moreover, when comparing with the evolution of bent Boolean functions, we notice that the difficulty for evolutionary algorithms is rather similar. Finally, we also tried evolving secondary constructions for self-dual bent functions, but this direction provided no successful results.
\end{abstract}


\section{Introduction}
\label{sec:intro}

Bent Boolean functions are interesting mathematical objects. For instance, they are used in coding theory with Kerdock codes~\cite{KERDOCK1972182} and to build bent function sequences for telecommunications~\cite{1056589}. Next, they are related to Golay complementary sequences and bent vectorial functions allow the construction of good codes. 
Bent Boolean functions are also often considered in cryptography since they achieve the highest possible nonlinearity values. Naturally, since bent Boolean functions are not balanced, they cannot be used directly in cryptographic algorithms but could be transformed into balanced Boolean functions. For instance, modified bent Boolean functions can also be used in block ciphers to create S-boxes, as in the case of the CAST-128 and CAST-256 ciphers~\cite{rfc2144}. 

Bent Boolean functions have been an important and active research domain for almost 50 years~\cite{Rothaus}.
As such, there are many works that consider how to construct bent Boolean functions.
The first and the most established approach is to use algebraic constructions. When considering algebraic constructions, it is common to divide them into primary and secondary constructions.
Primary constructions construct new functions from scratch by leveraging other types of mathematical objects such as permutations and partial spreads~\cite{dillon74,MCFARLAND19731}. Secondary constructions define new functions by starting from existing ones as building blocks~\cite{Rothaus}.
A second approach is to perform numerical simulations. There, one would commonly resort to random search or a certain kind of metaheuristics~\cite{Djurasevic2023}. Each of those approaches has specific advantages and drawbacks.

Metaheuristics represent an interesting approach for the construction of bent Boolean functions as they provide many different function instances and can work for different Boolean function sizes. Unfortunately, when going to larger sizes of Boolean functions, problems arise as there are $2^{2^n}$ Boolean functions of $n$ variables, and depending on the solution encoding, finding bent Boolean functions can become prohibitively difficult. For instance, one needs $2^n$ bits to encode a Boolean function of $n$ variables under a bitstring encoding. On the other hand, these difficulties also lead to considering the construction of bent Boolean functions as a benchmark problem~\cite{10.1145/3067695.3082535}. Consequently, we reach the situation where constructing bent Boolean functions becomes interesting for both application-driven reasons and benchmarking. 

The interest of the research community is evident from the plethora of works published every year on bent Boolean functions, see, e.g.,~\cite{carlet_2021,Mesnager2016,carlet16}.
However, we then reach an interesting problem. The research community made significant progress in the construction of bent Boolean functions, and the problem can hardly be considered difficult anymore. Indeed, Hrbacek and Dvorak used Cartesian Genetic Programming with various parallelization techniques to evolve bent Boolean functions up to 16 variables~\cite{10.1007/978-3-319-10762-2_41}. Picek and Jakobovic, on the other hand, used genetic programming to evolve algebraic constructions of bent Boolean functions~\cite{10.1145/2908812.2908915}. Both approaches can provide bent functions of many variables.

In this paper, we focus on a specific subclass called (anti-)self-dual bent Boolean functions. Such functions are much rarer than bent Boolean functions, satisfying the future requirements for benchmarking. Additionally, while the class of bent Boolean functions is small compared to the class of all Boolean functions, it is still large enough to make enumeration and classification impossible when $n\geq 10$~\cite{10.1504/IJICOT.2010.032864}. Thus, it makes sense to look for subclasses that are more constrained to generate and classify, satisfying also the requirements for practical relevance.
We note that the concept of self-dual bent functions is not new and has been discussed already in the 70s~\cite{Rothaus}, when Rothaus observed that ``many'' bent functions are equal to their duals, i.e., they are self-dual bent functions.

\subsection{Related Work}
To the best of our knowledge, there are no works that consider metaheuristics for the construction of (anti-)self-dual bent Boolean functions. On the other hand, there are many works that focus on generic bent functions.
Hrbacek and Dvorak used CGP to evolve bent Boolean functions up to 16 inputs~\cite{10.1007/978-3-319-10762-2_41}. The authors investigated various configurations
of algorithms to speed up the evolution process and succeeded in finding bent functions for sizes between 6 and 16 inputs.
Mariot and Leporati used a genetic algorithm to evolve semi-bent Boolean functions by spectral inversion~\cite{10.1007/978-3-319-26841-5_3}. The novelty of the approach is in using the Walsh-Hadamard spectrum as the genotype instead of the usual truth table-based bitstring encoding.
Picek and Jakobovic used GP to evolve algebraic constructions that are then used to construct bent Boolean functions~\cite{10.1145/2908812.2908915}. The authors presented results of up to 24 variables. On a similar research line, Mariot et al.~\cite{MariotSLM22} used evolutionary strategies to evolve a secondary construction based on cellular automata for quadratic bent functions.

Husa and Dobai used linear genetic programming to evolve bent Boolean functions~\cite{10.1145/3067695.3084220}. The authors reported better results than related works, and they managed to evolve bent Boolean functions of up to 24 inputs.
Picek, Sisejkovic, and Jakobovic investigated immunological algorithms to construct either bent or balanced, highly nonlinear Boolean functions~\cite{PICEK2017320}.
Picek et al. considered evolving quaternary bent Boolean functions, which are a generalization of bent (binary) Boolean functions~\cite{picek18a}.
Mariot et al. used evolutionary algorithms to evolve hyperbent Boolean functions~\cite{hyperbent}. Hyper-bent Boolean functions are a subclass of bent Boolean functions that also achieve maximum distance from all bijective monomial functions. Interestingly, the authors did not find such functions when using evolutionary algorithms.

We note that there is a large corpus of works considering metaheuristic techniques to construct balanced, highly nonlinear Boolean functions, see, e.g.,~\cite{10.1145/3579856.3590337,10.1145/3520304.3534087} but those works fall outside of our scope since duality is a concept that can be defined for bent Boolean functions only.

\subsection{Contributions}

This work focuses on the problem of evolving self-dual bent Boolean functions and conducts an extensive experimental evaluation. We consider both evolving (anti-)self-dual bent functions directly and evolving constructions of such functions.
Our main contributions are:
\begin{itemize}
    \item To the best of our knowledge, we are the first ones considering metaheuristics to evolve (anti-)self-dual bent Boolean functions. More precisely, we consider evolutionary algorithms and conduct experiments with two solution representations and two fitness functions. Our experiments with 8 to 16 inputs show we can successfully evolve such functions with the tree encoding.
    \item While (anti-)self-dual bent Boolean functions are rarer than general bent Boolean functions, we show that the difficulty for evolutionary algorithms is rather similar for both problems. Interestingly, when considering only nonlinearity, better results are obtained for anti-self-dual bent functions, while one would intuitively expect the same difficulty for self-dual and anti-self-dual bent functions. 
    \item Surprisingly, when evolving (anti)-self-dual bent functions in 16 variables, we managed to find such functions, while the corresponding search for the general bent functions gave no successful results in our experiments. This could indicate that for a large number of variables, constraining the search to a subclass can even be beneficial for metaheuristics. 
    \item We investigated the evolution of secondary constructions of (anti)-self-dual bent functions. The experiments did not result in any such construction, making it an open research question.
\end{itemize}

\section{Preliminaries}
\label{sec:background}
This section covers all necessary background information related to bent Boolean functions used throughout the paper. We start with some basic algebraic notation and then move to basic representations and properties of Boolean functions.

\subsection{Notation}

We denote by $\mathbb{F}_{2}$ the finite field with two elements, where sum and multiplication correspond respectively to the XOR (denoted as $\oplus$) and logical AND (denoted by concatenation) of the two operands. Given a positive integer $n \in \mathbb{N}^+$, the $n$-dimensional vector space over $\mathbb{F}_2$ is denoted as $\mathbb{F}_2^n$, while $\mathbb{F}_{2^n}$ is the finite field with $2^n$ elements. 
Up to isomorphism, there exists a unique field $\mathbb{F}_{2^n}$ of order $2^n$ for all $n \in \mathbb{N}$. Since this field is also an $n$-dimensional vector space, we can endow the vector space $\mathbb{F}_2^n$ with the structure of the field $\mathbb{F}_2^n$ when convenient. The usual inner product of $a,b \in \mathbb{F}_2^n$ equals $a\cdot b = \bigoplus_{i=1}^{n} a_{i}b_{i}$.

A Boolean function is any mapping $f: \F_2^n \to \F_2$ from $\mathbb{F}_{2}^{n}$ to $\mathbb{F}_{2}$, and it can be uniquely represented by its truth table, which is the list of pairs of function inputs (in $ \mathbb{F}_2^n$) and function values. 
The value vector is the binary vector composed of all $f(x), x \in \mathbb{F}_2^n$, where some total order has been fixed on $\mathbb{F}_2^n$ (most commonly, the lexicographic order). Since the size of the value vector equals $2^n$, the number of Boolean functions of $n$ variables is $2^{2^n}$, i.e., they grow super-exponentially in $n$. In practice, exhaustive enumeration of the set of $n$-variable Boolean functions becomes unfeasible already for $n>5$ (see Table~\ref{tab:nr_boolean}).

\begin{table}[t]
  \centering
  \caption{The number of Boolean functions.}
  \label{tab:nr_boolean}
  
  \begin{tabular}{ccccccccccc}
    &             \\
    $n$ & $3$ & $4$ & $5$ & $6$ & $7$ & $8$  & $9$ & $10$ & $11$ & $12$\\ \toprule
    $2^{2^n}$ & $2^8$ & $2^{16}$  & $2^{32}$ & $2^{64}$ & $2^{128}$ & $2^{256}$  & $2^{512}$ & $2^{1024}$ & $2^{2048}$ & $2^{4096}$\\\bottomrule
  \end{tabular}
\end{table}

The Walsh-Hadamard transform $W_{f}$ is another unique representation of a Boolean function that measures the correlation between $f(x)$ and the linear functions $a\cdot x$ (with the sum being calculated in ${\mathbb Z}$):
\begin{equation}
W_{f} (a) = \sum\limits_{x \in \mathbb{F}_{2}^{n}} (-1)^{f(x) \oplus a\cdot x}.
\end{equation}

The Walsh-Hadamard transform is very useful in cryptography, as many properties relevant to attacks on stream and block cipher models can be evaluated through it. To compute the Walsh-Hadamard transform efficiently, one can use the fast Walsh-Hadamard transform~\cite{carlet_2021}.

\subsection{Definitions, Properties, and Bounds}
\label{sec:boolean_properties}

A Boolean function $f$ is balanced if it takes the output value 1 exactly the same number $2^{n-1}$ of times as value 0 when the input ranges over ${\mathbb F}_2^n$.

The minimum Hamming distance between a Boolean function $f$ and all affine functions, i.e., the functions of algebraic degree at most 1 (of the same number of variables as $f$), is called the nonlinearity of $f$. The nonlinearity can be expressed in terms of the Walsh-Hadamard coefficients of $f$ as follows~\cite{carlet_2021}:
\begin{equation}
\label{eq:nonlinearity}
nl_{f} = 2^{n - 1} - \frac{1}{2}\max_{a \in \mathbb{F}_{2}^{n}} |W_{f}(a)|.
\end{equation}

Parseval relation states that the sum of the squared Walsh spectrum is constant for any Boolean function $f: \mathbb{F}_2^n \to \mathbb{F}_2$:
\begin{equation}
    \label{eq:parseval}
    \sum\limits_{a\in {\mathbb F}_2^n}W_f(a)^2=2^{2n}.
\end{equation}

Eq.~\ref{eq:parseval} implies that the nonlinearity of any $n$-variable Boolean function is bounded above by the so-called covering radius bound:
\begin{equation}
\label{eq_boolean_covering}
    nl_{f} \leq 2^{n-1}-2^{\frac{n}{2}-1}.
\end{equation}

Boolean functions can satisfy the bound~\ref{eq:parseval} with equality only if $W_f(a) = \pm 2^{\frac{n}{2}}$ for all $a \in \mathbb{F}_2^n$. Such functions are also called \emph{bent}, and they reach the maximum possible nonlinearity value $2^{n-1}-2^{n/2-1}$. Remark that bent Boolean functions exist only for $n$ even; see, for instance,~\cite{carlet_2021}.

For a bent function $f$ on $\mathbb F_{2^n}$, we define its \emph{dual} as the Boolean function $\widetilde{f}: \mathbb F_{2^n} \to \mathbb{F}_2$ satisfying:
\begin{equation}
\label{eq:self-dual}
    2^\frac{n}{2} (-1)^{ \widetilde{f}(x)}=W_f(x) \ \text{for all} \ x \in \mathbb F_{2^n}.
\end{equation}

The dual $\widetilde{f}$ of a bent function is also bent. A bent function $f$ is said to be self-dual if $\widetilde{f}(x) \oplus f(x) = 0$ for all $x \in \mathbb{F}_2^n$, and anti-self-dual if $\widetilde{f}(x) \oplus f(x) = 1$. Stated differently, a bent function is called self-dual if it is equal to its dual and anti-self-dual if it is equal to the complement of its dual.

Bent Boolean functions are rare, and we know the exact numbers of bent Boolean functions for $n\leq 8$ only. Self-dual bent functions are even rarer. We provide the numbers of bent and self-dual bent functions in Table~\ref{tab:nr_bent_dual}. Note that for the self-dual bent functions of 8 variables, we have results for quadratic functions (those with the algebraic degree at most two) only. The total number of self-dual bent functions is thus larger. There are as many anti-self-dual bent functions as there are self-dual bent functions.

\begin{table}[t]
\setlength{\tabcolsep}{6pt}
\centering
\caption{Number of bent and self-dual bent Boolean functions.}
\label{tab:nr_bent_dual}
\centering
\begin{tabular}{lcccc}\toprule
$n$      & 2 & 4   & 6                & 8      \\\midrule
\#bent & 8 & 896 & 5425430528 & $2^{106.3}$\\\midrule
\#self-dual & 2 & 20 & 42896 & 104960\\\bottomrule
\end{tabular}
\end{table}

For further information about bent Boolean functions and their properties, we refer interested readers to~\cite{MacWilliams-Sloane,carlet_2021}.

\section{Experimental Setup}
\label{sec:setup}

In this section, we discuss the solution representations, fitness functions, and evolutionary algorithm parameters.

\subsection{Solution Representations}

\paragraph{Bitstring encoding.} When evolving Boolean functions, the most common approach to encoding a solution is the bitstring representation, as discussed in~\cite{Djurasevic2023}. The bitstring represents the truth table of the function upon which the algorithm operates directly. For a Boolean function of $n$ variables, the truth table is encoded as a bitstring of length $2^n$. With this representation, the corresponding variation operators we use are the simple bit mutation, which inverts a randomly selected bit, and a mixing mutation, which shuffles the bits within a randomly selected substring.
For the crossover operators, we use the one-point crossover, which combines a new solution, taking the first portion from one and the second from the other parent, with a randomly selected breakpoint.
The second operator is the uniform crossover, which randomly selects a bit to be copied from both parents at each position into the child bitstring with uniform probability. Each time a crossover or mutation operation is invoked by the evolutionary algorithm, a random operator is chosen among the described ones.

\paragraph{Tree encoding.} The second approach in our experiments uses tree-based GP to represent a Boolean function in its symbolic form.
GP and its variants (such as Cartesian Genetic Programming~\cite{10.5555/2934046.2934074}) have already been extensively used in the evolution of Boolean functions and have been able to produce human-competitive results~\cite{Djurasevic2023,10.1145/3520304.3534087}.
In this case, we represent a candidate solution by a tree whose leaf nodes correspond to the input variables $x_1,\cdots, x_n \in \F_2$. The internal nodes are Boolean operators that combine the inputs received from their children and propagate their output to the respective parent nodes. 

Our experiments use the NOT function that takes a single argument, the function set operating on two arguments: OR, XOR, AND, AND2\footnote{The function AND2 behaves the same as the function AND but with the second input inverted.}, and XNOR, and the function IF, which takes three arguments and returns the second one if the first one evaluates to true and the third one otherwise.
The output of the root node is the output value of the Boolean function. The corresponding truth table of the function $f: \F_2^n \to \F_2$ is determined by evaluating the tree over all possible $2^n$ assignments of the input variables at the leaf nodes. 
Each GP individual is evaluated according to the truth table it produces.
The genetic operators used for GP are simple tree, uniform, size fair, one-point, and context preserving crossover~\cite{poli08:fieldguide} (selected at random each time crossover is performed), and subtree mutation.

Since the search size grows rapidly with the number of inputs, we expect the bitstring encoding to perform much worse than the GP encoding, which is in accordance with most of the previous works, as discussed before.
However, we include both encodings for completeness and a more reliable estimate of the problem's difficulty.

\subsection{Booolean Function Constructions}

Rather than evolving a Boolean function with the desired properties directly, it is possible to define one by combining existing functions, following the secondary construction approach~\cite{10.1145/2908812.2908915}. When leveraging metaheuristics such as GP, this is usually done by using predefined Boolean functions (seed functions) of a smaller number of inputs, with the addition of the remaining inputs (up to $n$) as independent variables.
Since bent functions only exist when the number of variables is even, the minimal incremental step would include seed functions with $n$ variables and the addition of two independent variables to obtain a function of $n+2$ variables. For instance, one may have several seed functions of $n=4$ variables (available in truth table form) and combine them with two additional variables to form a new $6$-variable function.
An example of this type of construction evolved by GP for balanced, highly nonlinear Boolean functions, taken from~\cite{CarletDJMP22}, is shown in Figure~\ref{fig:tree-cons}.

\begin{figure}[t]
    \centering
    \begin{minipage}{.4\columnwidth}
    \begin{tikzpicture}[scale=0.85]
    \tikzset{every tree node/.style={draw, thick, circle,inner sep=1pt, minimum size=0.7cm}}
    \tikzset{every leaf node/.append style={fill=black!10}}
    \tikzset{edge from parent/.append style={thick}}
        \Tree [.IF $x_0$ $f_0$ [ .XOR $x_1$ $f_1$ ]
        ]
    \end{tikzpicture}
    \end{minipage}%
    \begin{minipage}{.6\columnwidth}
    \begin{equation}
    \label{eq:conc-cons}
    F(x_0,x_1,x) = 
    \begin{cases}
        f_0(x) \enspace , & \textrm{ if } x_0 = 1 \enspace , \\
        f_1(x) \oplus x_1 \enspace, & \textrm{ if } x_0 = 0 \enspace .
    \end{cases}
    \end{equation}
    \end{minipage}
    \caption{An example of secondary construction evolved by GP.}
    \label{fig:tree-cons}
\end{figure}

To evolve constructions with GP, one requires the existence of a certain number of seed bent functions that are included in the terminal set.
In our experiments, up to four predefined Boolean functions are available as terminals, denoted as $f_0$, $f_1$, $f_2$, and $f_3$.
The number of variables for the seed functions is taken to be $n$, and they are encoded by their truth tables.
Additionally, the terminal set includes two additional Boolean variables, $x_0$ and $x_1$. 
Consequently, the resulting construction (in the form of a GP tree) represents a new Boolean function with $n + 2$ variables.
The function and the operator set remain the same as in the direct search GP strategy. 

To apply this approach, the seed functions must either be given or previously evolved.
The initial set of seed functions is obtained with direct evolution, starting with a low number of variables (e.g., four or six variables), which is trivial to find.
Then, the seed functions are used to find constructions for a larger number of variables.

To obtain a \textit{general} construction, rather than an expression that works only on a single set of seed functions, we employ multiple sets of seed functions and evaluate the same construction on all of them. 
Here, we employ two evaluation schemes:
\begin{itemize}
    \item \textbf{incremental}, in which the perfect score (however it may be defined) must first be obtained on the first group of seed functions, and only then the construction is further evaluated on remaining groups to promote generality;
    \item \textbf{concurrent}, in which all seed groups are used concurrently to evaluate the construction. 
\end{itemize}
In both schemes, the total score is simply the sum of individual scores on all seed groups that were used.

\subsection{Fitness Functions}
\label{subsec:fit}

To evolve bent Boolean functions, one only needs to check that the maximal absolute value in the  Walsh-Hadamard transform equals $2^{\frac{n}{2}}$ (see Eq.~\eqref{eq:nonlinearity}).
For self-dual functions, each Walsh-Hadamard coefficient must not only be equal to this absolute value: considering Eq.~\eqref{eq:self-dual}, its sign must agree with the corresponding output value in the function's truth table. For instance, if $f(a) = 0$ for $a \in \mathbb{F}_2^n$, the corresponding coefficient $W_f(a)$ in the Walsh-Hadamard transform must assume the value of $+2^{\frac{n}{2}}$, and $-2^{\frac{n}{2}}$ otherwise; for anti-self-dual functions, the previous values are inverted.

The remark above suggests the following strategy for our first fitness function (denoted as $fit_1$): count the number of entries in the Walsh-Hadamard transform whose absolute value is equal to $2^{\frac{n}{2}}$ and, at the same time, the sign of the value matches the corresponding output value in the truth table. Formally, given $f: \mathbb{F}_2^n \to \mathbb{F}_2$, its fitness score under $fit_1$ is defined as:
\begin{equation}
    \label{eq:fit1}
    fit_1(f) = |\{a \in \mathbb{F}_2^{n}: \ W_f(a) = 2^{\frac{n}{2}}\cdot (-1)^{f(a)} \}| \enspace .
\end{equation}
Since the number of entries in the Walsh-Hadamard transform is equal to the truth table size ($2^n$), the range of this fitness function is $[0, \dots , 2^n]$, where $2^n$ denotes the optimal value that corresponds to a self-dual bent function.
Note that this fitness function will drive the search towards the bent and self-dual criteria at the same time, as opposed to a lexicographic approach that would first try to achieve a bent function and then additionally optimize duality.

The second fitness function we employ takes a closer look into the deviation of each Walsh-Hadamard entry from the desired value.
Apart from the number of correct values, as evaluated by $fit_1$, we sum the absolute differences (from either $2^{\frac{n}{2}}$ or $-2^{\frac{n}{2}}$) of every incorrect coefficient, and divide the sum with the product of the maximal possible difference ($2^{\frac{n}{2}}$) by the total number of entries ($2^n$).
Consequently, the deviation part is normalized in $[0,1]$, and its difference from 1 is simply added to the number of correct entries computed through $fit_1$. Hence, the fitness score of $f: \mathbb{F}_2^n \to \mathbb{F}_2$ under $fit_2$ is formally defined as:

\begin{equation}
    \label{eq:fit2}
    fit_2(f) = fit_1(f) + \left[ 1 - \frac{\sum_{a \in \mathbb{F}_2^n} \left| 2^{\frac{n}{2}}\cdot (-1)^{f(a)} - W_f(a)\right|}{2^n \cdot 2^{\frac{n}{2}}} \right] \enspace .
\end{equation}

The integer part of $fit_2$ always equals the value obtained with $fit_1$. In particular, when the normalized sum of the deviations is 0 (that is, we reached an optimal solution), the difference from 1 is not added to $fit_1$. Thus, the optimal fitness value for $fit_2$ is the same as $fit_1$, i.e., $2^n$.

When trying to evolve secondary constructions of dual-bent functions, we use the same two fitness functions. However, the fitness is evaluated separately for each set of seed functions. 
In our experiments, we use \textit{four} independent sets to promote (although not guarantee) generality.
With these settings, the optimal fitness value for a construction would equal four times the maximal value ($2^n$) for the defined criteria.

Finally, since (anti-)self-dual functions are a subset of bent functions, we also included experiments that optimize only the nonlinearity property, trying to obtain bent functions.
In this experiment, only the tree-based GP representation is used since this approach has shown to be the most efficient one in evolving bent Boolean functions~\cite{Djurasevic2023}.

\section{Experimental Results}
\label{sec:results}

In this section, we first describe the settings adopted for our experimental evaluation of the evolutionary algorithms described in the previous section to design self-dual bent functions. Then, we report the results of the experiments for the direct search and the evolution of secondary construction approaches.

\subsection{Experimental Settings}

Both bitstring and GP encoding employed the same evolutionary algorithm: a steady-state selection scheme with a 3-tournament elimination operator. 
In each iteration of the algorithm, three individuals are chosen at random from the population for the tournament, and the worst one in terms of fitness value is eliminated. 
The two remaining individuals in the tournament are used by the crossover operator to generate a new child individual, which then undergoes mutation with individual mutation probability $p_{mut} = 0.5$. Finally, the mutated child takes the place of the eliminated individual in the population.

The population size in all experiments was 500, and the termination criteria were set to $10^6$ evaluations. Finally, each experiment was repeated $30$ times.
The maximum tree depth for the GP representation was based on a set of preliminary experiments and was set to $\min(5, n-5)$, where $n$ is the number of Boolean variables (which also represents the size of the terminal set).

\subsection{Directly Evolving (Anti)-Self-Dual Bent Boolean Functions}

The results obtained when directly evolving self-dual bent Boolean functions are outlined in Table~\ref{tab:best}. 
The results demonstrate that in all cases when using GP, we can find the target (anti-) dual function in at least one run, which is evident from the fact that the best fitness value was equal to $2^n$, meaning that every WH entry was correct.
Furthermore, for $n<10$, GP found a self-dual or anti-self-dual function in every run, demonstrating the effectiveness of this approach. 

As expected, the bitstring representation (denoted with TT) obtained inferior results when compared to GP and could not reach the target values, i.e., it did not find a self-dual or anti-self-dual function in even one of the runs (for this reason, we omit the results for anti-self-dual bent functions). 
When observing the influence of the fitness function variant used in the experiments, we see no difference between the two fitness functions when using the GP representation, as in both cases, the target functions were obtained. 
On the other hand, for the TT representation, there are some minor differences, although not consistently in favor of a single fitness function.
\begin{table}
\caption{Best obtained fitness values when optimizing for self-dual and anti-self-dual bent Boolean functions.}
\label{tab:best}
\setlength{\tabcolsep}{6pt}
\centering
\begin{tabular}{@{}lcrrrrrr@{}}
\toprule
\multirow{2}{*}{Representation} & \multirow{2}{*}{Fitness function} & \multicolumn{1}{l}{}  & \multicolumn{5}{c}{Variables}                                                                                             \\ \cmidrule(l){3-8} 
                              &  & \multicolumn{1}{l}{6} & \multicolumn{1}{c}{8} & \multicolumn{1}{c}{10} & \multicolumn{1}{c}{12} & \multicolumn{1}{c}{14} & \multicolumn{1}{c}{16} \\ \midrule
self-dual  TT & $fit_1$                   & 41                    & 69                    & 101                    & 169                    & 257                    & 471                    \\
self-dual TT & $fit_2$                   & 43                    & 70                    & 103                    & 168                    & 256                    & 481                    \\ \midrule
self-dual GP & $fit_1$                      & 64                    & 256                   & 1024                   & 4096                   & 16384                   & 65536                  \\
self-dual GP & $fit_2$                      & 64                    & 256                   & 1024                   & 4096                   & 16384                  & 65536                  \\ \midrule
anti-self-dual GP & $fit_1$                 & 64                    & 256                   & 1024                   & 4096                   & 16384                   & 65536                  \\
anti-self-dual GP & $fit_2$                 & 64                    & 256                   & 1024                   & 4096                   & 16384                  & 65536                  \\ \bottomrule
\end{tabular}
\end{table}

\begin{figure}
    \centering
    \includegraphics[trim={2.95cm 2.95cm 0 0},clip,width=0.83\linewidth]{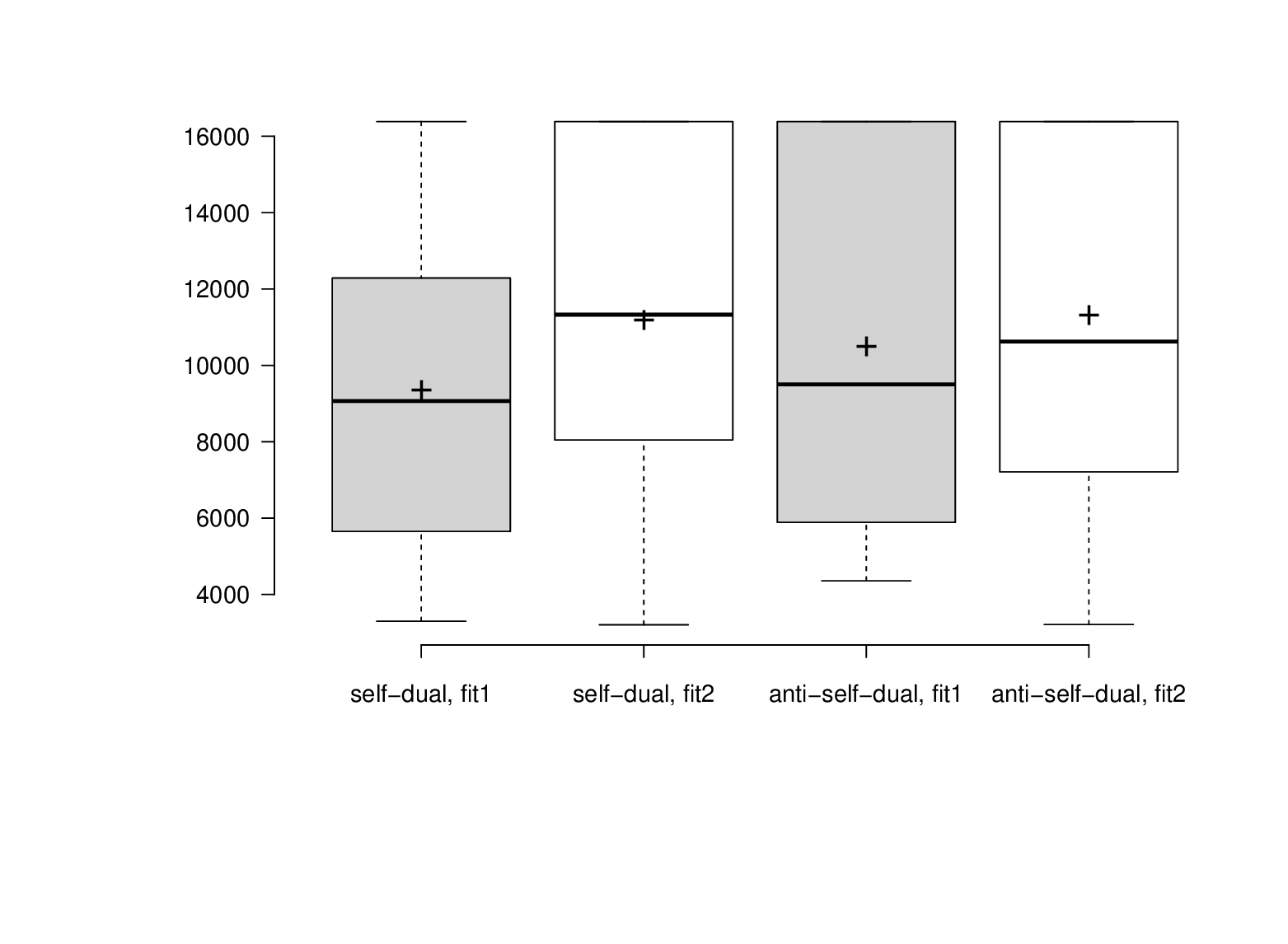}
    \caption{Box plots of the results obtained when optimizing for self-dual and anti-self-dual bent Boolean functions considering functions of 14 variables.}
    \label{fig:dual_14}
\end{figure}

\begin{figure}
    \centering
    \includegraphics[trim={2.95cm 2.95cm 0 0},clip,width=0.83\linewidth]{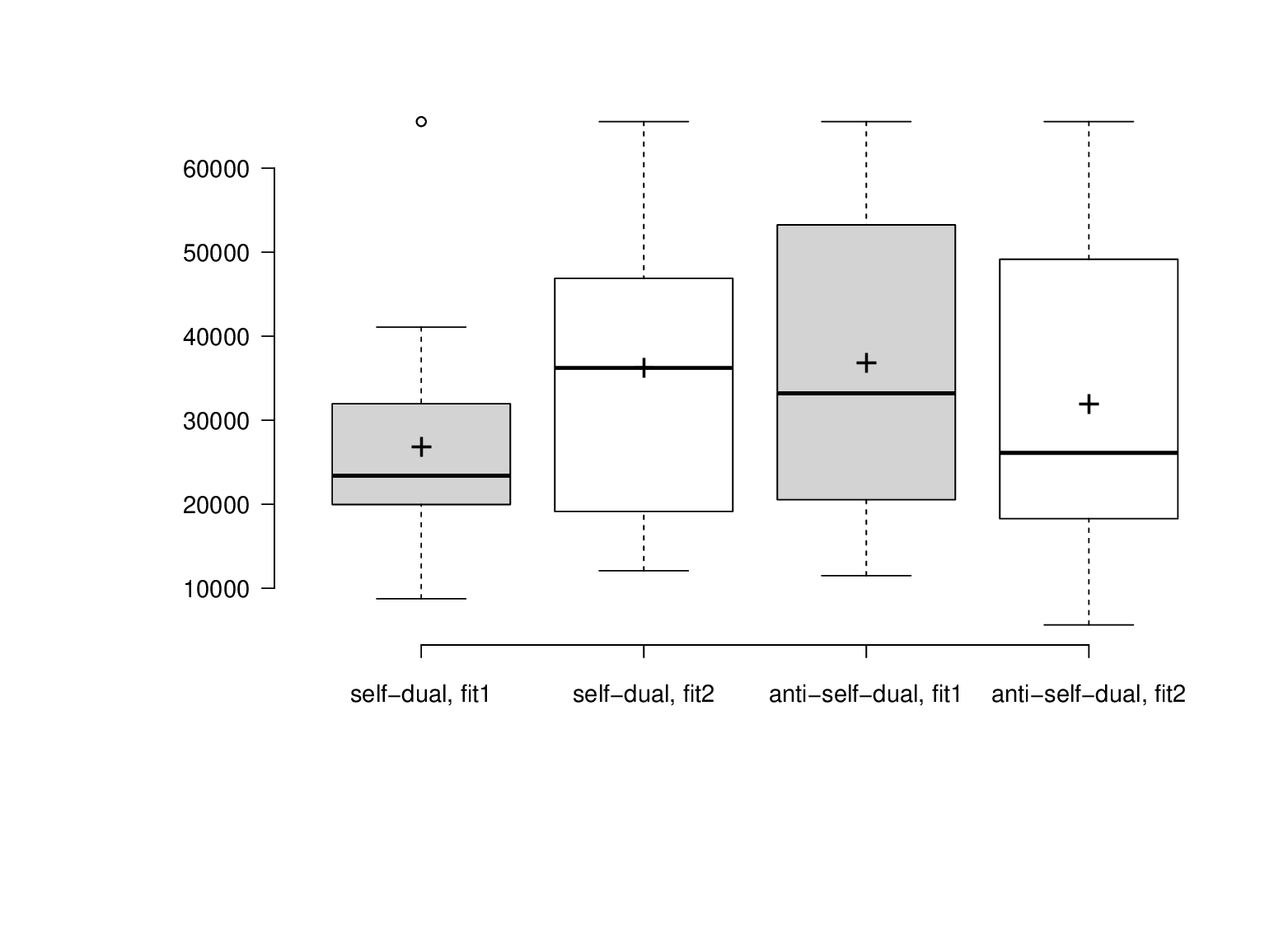}
    \caption{Box plots of the results obtained when optimizing for self-dual and anti-self-dual bent Boolean functions considering functions of 16 variables.}
    \label{fig:dual_16}
\end{figure}

The results are additionally presented as box plots to better outline the distribution obtained across all executions of the GP algorithm (TT is not considered here due to its inferior performance). 
Figure~\ref{fig:dual_14} provides the results obtained by GP for Boolean functions considering 14 variables.
The figure shows that $fit_2$ leads to somewhat better results in both cases as the median values obtained by it are slightly higher.
However, Figure~\ref{fig:dual_16} demonstrates that for 16 variables, neither fitness function definition consistently achieved a better performance.
Therefore, it is impossible to conclude that either fitness function definition is significantly better. Still, we recommend the second fitness since it provides more information and can potentially direct the search better.

\begin{figure}
    \centering
    \includegraphics[trim={2.95cm 2.95cm 0 0},clip,width=0.83\linewidth]{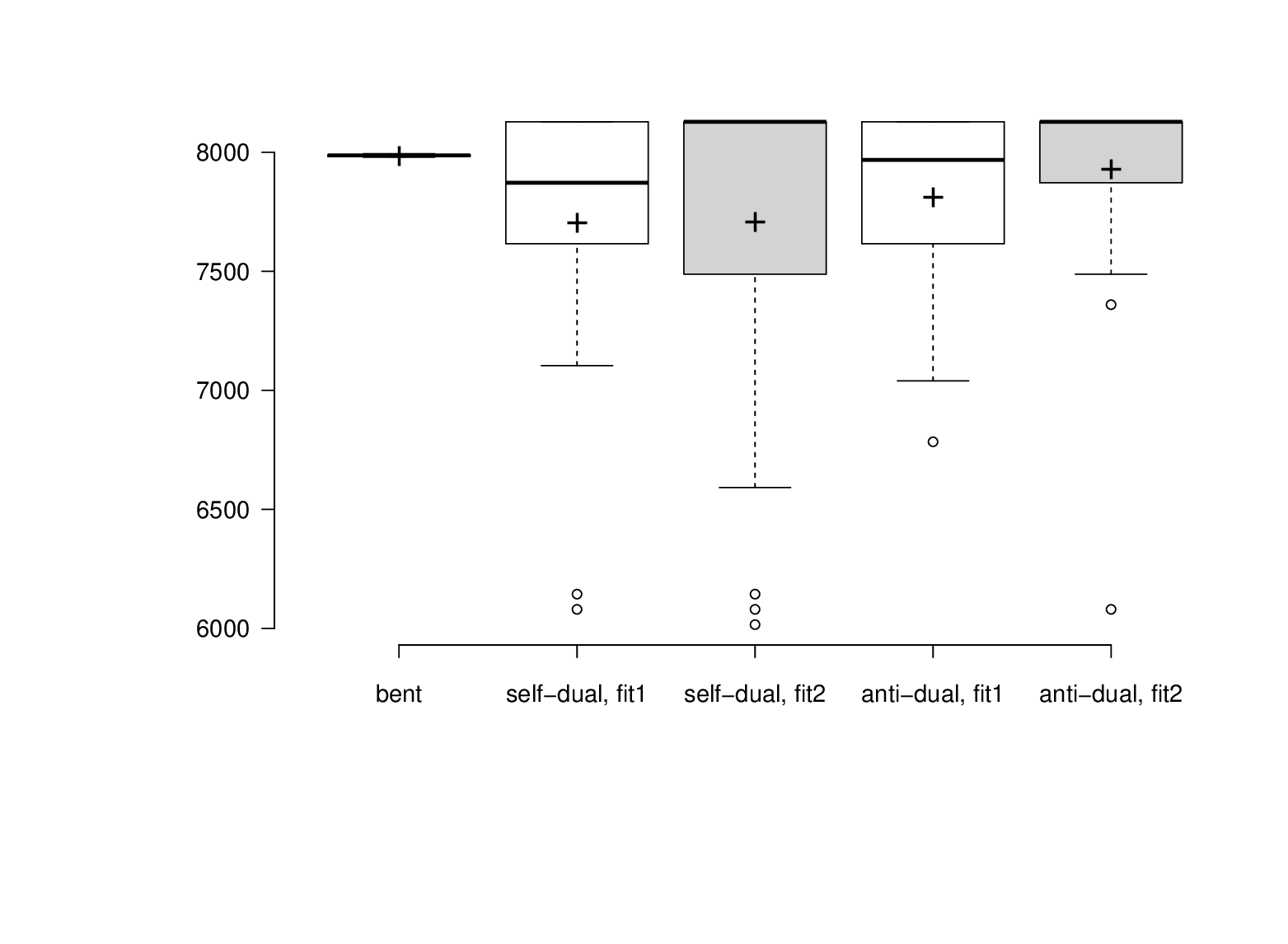}
    \caption{Achieved nonlinearity levels - 14 variables.}
    \label{fig:nl_14}
\end{figure}

\begin{figure}
    \centering
    \includegraphics[trim={2.95cm 2.95cm 0 0},clip,width=0.83\linewidth]{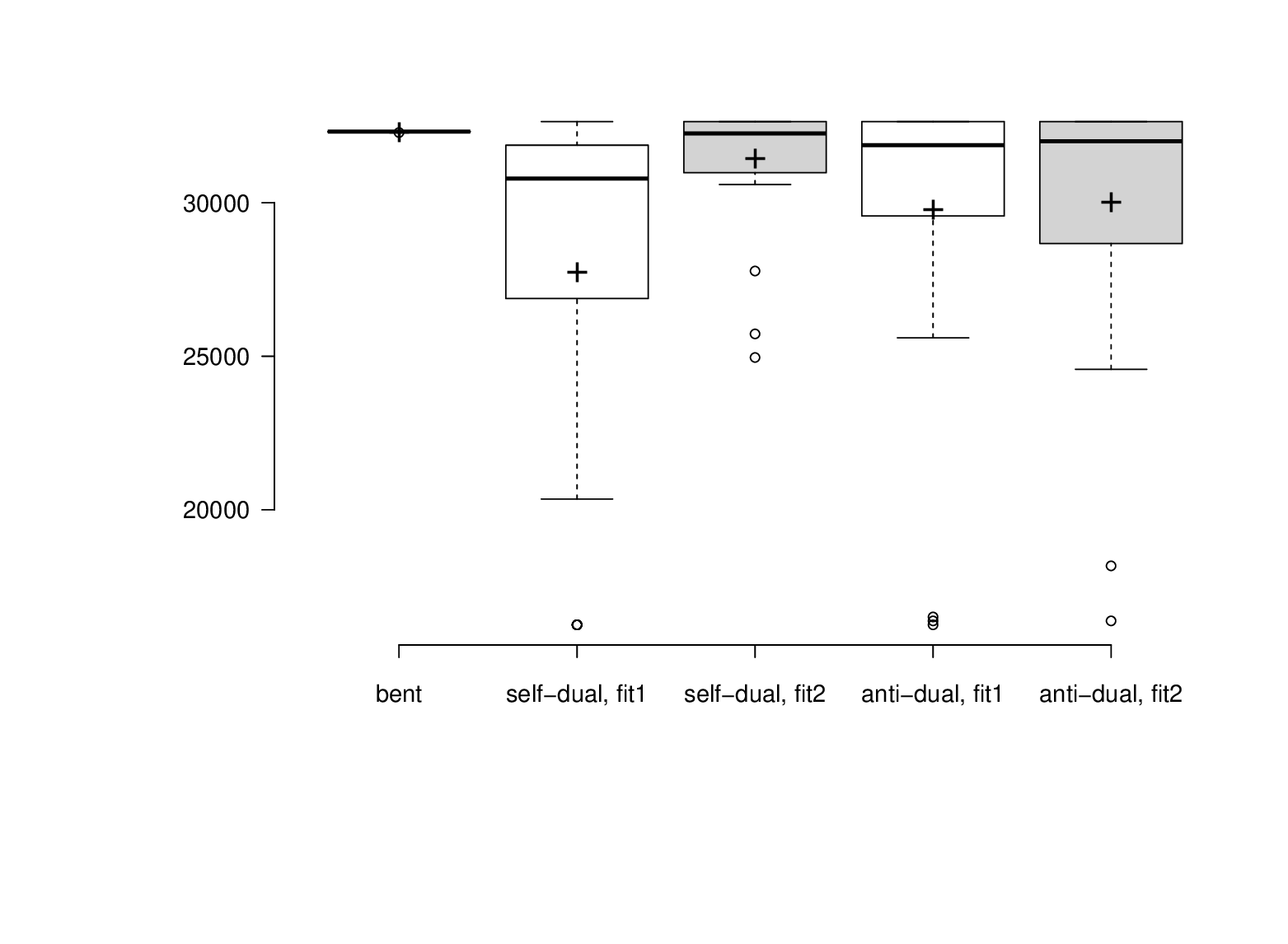}
    \caption{Achieved nonlinearity levels - 16 variables.}
    \label{fig:nl_16}
\end{figure}

Figures~\ref{fig:nl_14} and~\ref{fig:nl_16} show the nonlinearity levels of the obtained Boolean functions considering 14 and 16 variables, respectively. 
In this case, we compare the results obtained when optimizing only for nonlinearity, denoted in the figure under the column 'bent', against those obtained by optimizing for self-dual or anti-self-dual functions. 
First, we can notice from the results that when optimizing directly for nonlinearity, the results' dispersion is rather small, and the algorithm obtains similar results across all runs. 
In the remaining cases, the results are significantly more dispersed, and often, poor nonlinearity levels are obtained.
However, such behavior is expected as, in these cases, the nonlinearity was not optimized directly, and therefore, it is expected that functions with poor nonlinearity may be found.

When considering the nonlinearity levels, it seems that $fit_2$ more often leads to better results, meaning it might be a better choice in this case. 
Regarding self-dual and anti-self-dual optimization, the results also demonstrate that better nonlinearity values were obtained when optimizing for anti-self-dual functions. However, more experimental runs should be conducted to verify if this effect is statistically significant.
An additional interesting observation from the results is that when 16 variables were considered, optimizing for self-dual or anti-self-dual functions always resulted in at least one bent - and at the same time dual - function (with a nonlinearity level equal to 32640), which is evident from Table~\ref{tab:best}.
On the other hand, by optimizing directly for nonlinearity, we did not obtain functions with this nonlinearity value. 

\subsection{Evolving Secondary Constructions of Self-Dual Bent Boolean Functions}

Table~\ref{tab:secondary} outlines the results obtained when evolving constructions of self-dual bent Boolean functions. 
The results demonstrate that constructions from 4 to 6 failed as the target score is 256 since we have 4 test constructions (4 seed sets), each of which should have a value of 64. 
In the incremental construction scheme, obtaining the target score for the first seed was possible but not for the three remaining seed sets as well. 
Nevertheless, we opted to also test the secondary constructions from 6 to 8 variables since the number of (anti-self)-dual bent functions for 4 inputs is only 20 (see Table~\ref{tab:nr_boolean}), making it potentially too small a pool size.
Similar results were also obtained for the construction from 6 to 8 variables; in that case, the target score would equal 1024. However, in neither of the experiments did the algorithm come even close to that value.
Even worse, for this construction experiment, the algorithm could not find even a single self-dual construction for any given seed set. 
Finally, we mention that we also tested the secondary constructions where we seeded with the general bent functions, but the experiments were not successful.
We emphasize that our approach did find ``constructions'' (e.g., $x_0x_1 \oplus f_0$, which is just a concatenation of one seed function and its inverse), but we excluded such constructions from considerations since they are trivial.

\begin{table}
\caption{Best results for secondary constructions}
\label{tab:secondary}
\setlength{\tabcolsep}{6pt}
\centering
\begin{tabular}{@{}lllll@{}}
\toprule
Input size & Target size & Evaluation scheme       & Fitness 1 & Fitness 2 \\ \midrule
4          & 6           & incremental & 194     & 203.719   \\
4          & 6           & concurrent       & 238       & 238.906   \\
6          & 8           & incremental & 192       & 192.75    \\ 
6          & 8           & concurrent       & 792       & 795.906   \\ \bottomrule
\end{tabular}
\end{table}

We note there are both primary and secondary constructions of (anti)-self-dual bent functions known~\cite{10.1504/IJICOT.2010.032864}, which means our algorithm failed, and not that such constructions are impossible. Moreover, the known secondary construction is rather simple from the construction perspective, requiring only XOR and inner product (indirect sum construction), but the seed functions need to fulfill specific requirements. 

\section{Conclusions and Future Work}
\label{sec:conclusions}

This paper tackles the problem of evolving (anti)-self-dual bent functions. Our results show that the tree encoding is much more efficient than the bitstring (truth table-based) one. This is aligned with the related works and results on general bent Boolean functions. We can observe that the problem also does not seem much more difficult than evolving general bent functions since GP manages to achieve it for every dimension. Interestingly, we also observe that when evolving for nonlinearity only, better results are obtained for anti-self-dual bent functions, which (intuitively) should not be easier than considering self-dual bent functions.
On the other hand, the construction approach did not succeed, making it an interesting future research direction to understand why. One option could be to allow the seed functions to have different dimensions.
Next, one could also consider evolving primary constructions of (anti-)self-dual bent functions, in which case, no seed functions are required.
Finally, since the problem of evolving (anti-)self-dual bent functions is not difficult enough (or, more precisely, is still simpler than we expected), it would be interesting to consider even smaller subsets, like if there are (anti)-self-dual bent functions that are also rotation symmetric (invariant under cyclic shift). To the best of our knowledge, this is also something that is not known in general.

\bibliographystyle{abbrv}
\bibliography{bibliography}

\end{document}